# nnY-Net: Swin-NeXt with Cross-Attention for 3D Medical Images Segmentation

Haixu Liu[1], Zerui Tao[1], Wenzhen Dong[2], and Qiuzhuang Sun[1]

[1]The University of Sydney; [2]The Chinese University of Hong-Kong
{hliu2490,ztao0063}@uni.sydney.edu.au; dongwz@link.cuhk.edu.hk; qiuzhuang.sun@sydney.edu.au

## I. Abstract

This paper provides a novel 3D medical image segmentation model structure called nnY-Net. This name comes from the fact that our model adds a cross-attention module at the bottom of the U-net structure to form a Y structure. We integrate the advantages of the two latest SOTA models, MedNeXt and SwinUNETR, and use Swin Transformer as the encoder and ConvNeXt as the decoder to innovatively design the Swin-NeXt structure. Our model uses the lowest-level feature map of the encoder as "Key" and "Value" and uses patient features such as pathology and treatment information as Query to calculate the attention weights in a Cross Attention module. Moreover, we simplify some pre- and post-processing as well as data enhancement methods in 3D image segmentation based on the dynUnet and nnU-net frameworks. We integrate our proposed Swin-NeXt with Cross-Attention framework into this framework. Last, we construct a DiceFocalCELoss to improve the training efficiency for the uneven data convergence of voxel classification.

## II. Introduction and Related work

Liver tumor CT 3D image segmentation is an important task in medical image segmentation. Accurate segmentation of CT images can help doctors estimate the volume of the tumor and develop a reasonable treatment plan. One mainstream solution to deal with this task before 2016 was the region growing method, an unsupervised algorithm. Çiçek et al. [1] proposed the 3D U-Net to extend the classical model on semantic segmentation of images to three dimensions. The nnU-Net [2] further improves upon U-Net and integrates automated pre-processing and post-processing techniques with parameter selection guidelines [2]. The good performance of nnU-Net makes it a common baseline for medical image segmentation tasks. Using a Transformer structure, UNETR utilizes ViT as an encoder and retains the convolutional layer of U-Net as a decoder [3]. SwinUNETR further uses a feature extraction network, Swin Transformer, to replace the ViT encoder in UNETR, and became the new SOTA on some segmentation tasks such as liver tumors [4]. Recently, [5] proposed MedNeXt based on the Conv-NeXt's U-Net structure and integrated the pre- and post-processing techniques of nnU-Net. The performance of this model is excellent on some segmentation tasks
.

## III. Exploratory Data Analysis

The data consists of two parts, one containing 110 liver CT scan images of 98 patients and the corresponding labels and the other containing the patients' pathology and treatment data.

First, we examined the patient pathology and treatment data. The raw data has 56 variables excluding the index columns, and we find that around 10 patients have missing RECIST, EASL, and Interval_FU metrics, 22 patients have missing pathology grading, and 52 patients have missing tumor size.

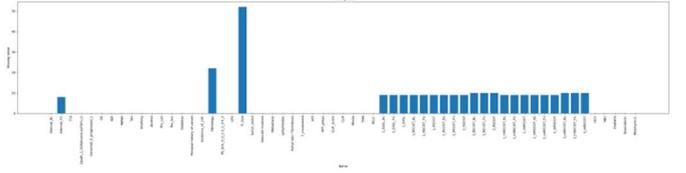

Fig. 1 Missing value statistics for each feature.

Next, we find that there are six labels of HCC 017, and there are inconsistencies between the label dimensions and image dimensions in HCC 008 and 025. We further checked the 3D images and found that some labels of HCC 009 do not correspond to the images. To avoid affecting the training process, we deleted the four groups of files mentioned above from the training set.

Then we checked the distribution of voxel values for the remaining images. Some parts of the images are labeled with -2048 for the non-human part, while some images are labeled -1024 for both the non-human part and air. We find that the background labels account for more than 90% of the pixels. Therefore, we crop 3D images to reduce the background as pre-processing.

For spatial resampling and Resize, we observed the size and spacing distributions in the three dimensions of the CT image. And plotted as follows:

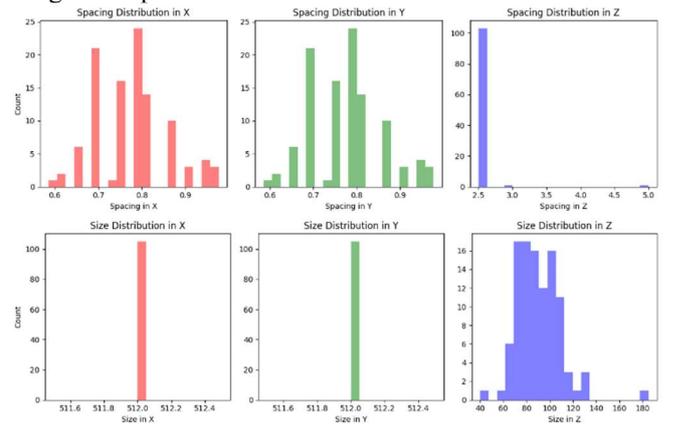

Fig. 2 Distribution of spacing and size.

We also explored the distribution of connected block sizes and the distribution of voxel values in each label.

## IV. Method

An overview of our method is shown in Fig. 3. We detail each module as follows.

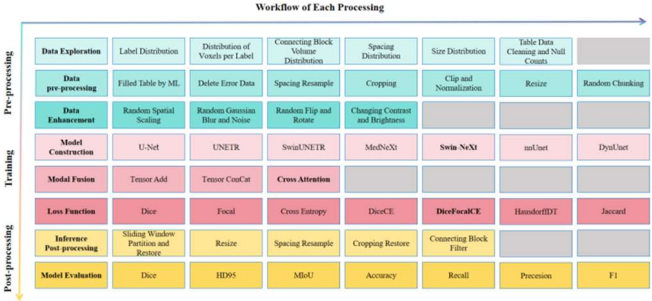

Fig. 3 Workflow for this article

Because the final work needs to be submitted through Jupyter Notebook, we develop our model based on a medical image processing package, MONAI, that is friendly to the Notebook code style.

### A. Data Preprocessing

Data preprocessing is mainly divided into two parts, (i) tabular data cleaning and filling and (ii) 3D image pre-processing and data enhancement. First, missing tabular data can be filled by using the data without missing labels to train machine learning models. Specifically, for a specific feature, we train 11 machine learning models based on all available data. The model with the best R-square (for continuous features) or F1 (for categorical features) is used for missing data imputation. We matched the processed tabular data with nrrd file paths and seg.nrrd file paths via dictionary for data loading.

Then we preprocess and enhance the 3D image by first cropping the images to remove background. We next perform Clip operation on the dataset based on the distribution of voxel values in the data exploration to remove the pixel values that are too high or too low. We perform normalization to enhance the contrast and select the median value of voxel spacing in each dimension for voxel resampling. The standard scale for image deflation was selected as the median value of the number of voxels on each dimension for all images. After that, we chunk the images and labels and specify the chunking scale to be half of the standard scale for each dimension to avoid exceeding the memory during training. These images are then subjected to some of the enhancement methods utilized in dynUnet such as spatial and contrast transformations, adding Gaussian blur and Gaussian noise, etc. Our preprocessing enhances the model's ability for convergence.

### B. Model Formulation and Training

We use Swin Transformer as the encoder and Conv-NeXt as the decoder to innovatively design a new network structure called Swin-NeXt. We briefly review these models as follows.

The input to the Swin Transformer model $\chi \in \mathbf{R}^{H \times W \times D \times S}$ is a token with a patch resolution of $(H', W', D')$ and dimension of $H' \times W' \times D' \times S$. We first utilize a patch partition layer to create a sequence of 3D tokens $\left\lceil \frac{H}{H'} \right\rceil \times \left\lceil \frac{W}{W'} \right\rceil \times \left\lceil \frac{D}{D'} \right\rceil$ and project them into an embedding space with dimension of dimension C. The self-attention is computed into non-overlapping windows that are created in the partitioning stage for efficient token interaction modeling. Specifically, we utilize windows of size $M \times M \times M$ to evenly partition a 3D token into $\left\lceil \frac{H'}{M} \right\rceil \times \left\lceil \frac{W'}{M} \right\rceil \times \left\lceil \frac{D'}{M} \right\rceil$ regions at a given layel l in the transformer encoder. Subsequently, in layerl + 1, the partitioned window regions are shifted by $\left( \left\lfloor \frac{M}{2} \right\rfloor, \left\lfloor \frac{M}{2} \right\rfloor, \left\lfloor \frac{M}{2} \right\rfloor \right)$ voxels. In layers l and l + 1 in the encoder, the outputs are calculated as

$$\hat{z}^l = W - MSA\left(LN(z^{l-1})\right) + z^{l-1}$$
$$z^l = MLP\left(LN(\hat{z}^l)\right) + \hat{z}^l$$
$$\hat{z}^{l+1} = SW - MSA\left(LN(z^l)\right) + z^l$$
$$z^l + 1 = MLP\left(LN(\hat{z}^{l+1})\right) + \hat{z}^{l+1}$$

Here, W-MSA and SW-MSA are regular and window partitioning multi-head self-attention modules respectively; $\hat{z}^l$ and $\hat{z}^{l+1}$ denote the outputs of W-MSA and SW-MSA; MLP and LN denote layer normalization and Multi-Layer Perceptron respectively. For efficient computation of the shifted window mechanism, we leverage a 3D cyclic shifting [24] and compute self-attention according to

$$\text{Attention}(Q, F, V) = \text{Soft max}\left(\frac{QK^T}{\sqrt{d}} + B\right)V$$

where Q, K, V denote queries, keys, and values respectively; d represents the size of the query and key, B is a relative position bias[6]. Fig. 4 graphically illustrates W-MSA and SW-MSA operations for 3D images.

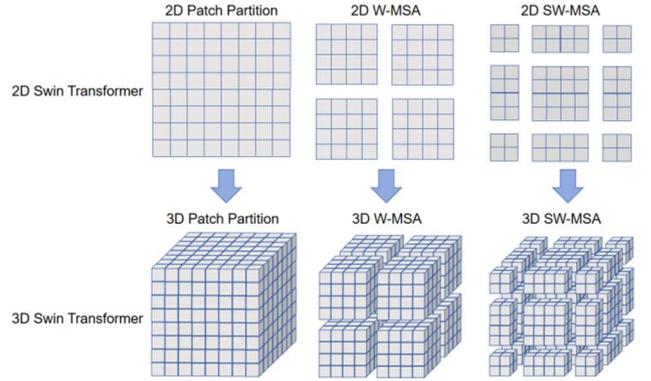

Fig. 4 Extending the Swin transformer from 2D to 3D

The decoder MedNeXt uses the following procedures. First, assuming the dimension of the input feature map X is $D \times H \times W$ (depth D, height H, width W), we first expand X to obtain $\tilde{X}$ by inserting $s - 1$ zeros between adjacent layers, rows, and columns (where s is the stride). This expansion increases the size of the spatial dimensions of the expanded X. We then apply the 3D convolutional kernel K (with dimensions $F_d \times F_h \times F_w$) to the expanded input $\tilde{X}$, resulting in the upsampled output feature map Y. In this step, the standard 3D convolution operation is applied to the expanded input. Each voxel of the output feature map Y can be calculated using the following formula:

$$Y_{i,j,k} = \sum_{d=0}^{F_d - 1} \sum_{m=0}^{F_h - 1} \sum_{n=0}^{F_w - 1} K_{d,m,n} \cdot \tilde{X}_{(i-d)s, (j-m)s, (k-n)s}$$

where i, j, k are the depth, row, and column indices of the voxel in the output feature map Y, respectively, $K_{d,m,n}$ is the element in the 3D convolutional kernel K, and $\tilde{X}_{(i-d)s, (j-m)s, (k-n)s}$ is the voxel in the expanded input $\tilde{X}$ that corresponds and multiplies with $K_{d,m,n}$.

MedNeXt also uses a Gaussian error linear unit (GELU) activation function[7], given by

$$\text{GELU}(x) = \frac{x}{2}\left[1 + \text{erf}\left(\frac{x}{\sqrt{2}}\right)\right]$$

where erf is the cumulative distribution function of a standard normal distribution.

$$\text{erf}(z) = \frac{2}{\sqrt{\pi}}\int_0^z e^{-t^2}dt$$

Combining these two functions, we can obtain the complete mathematical description of the GELU function:

$$\text{GELU}(x) = \frac{x}{2}\left[1 + \frac{2}{\sqrt{\pi}}\int_0^{\frac{x}{\sqrt{2}}} e^{-t^2}dt\right]$$

Based on the above description, the proposed model has the following structure shown in Fig. 5.

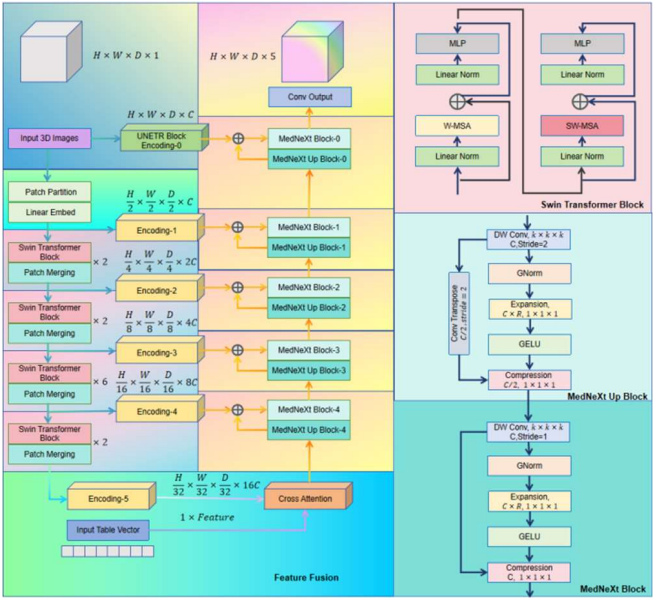

Fig. 5 The structure of the proposed model.

To include the information of tabular data for model training, we consider fusing the features of 3D images and tabular data in the last layer of the encoder, the Bottleneck layer. This can be done in three possible ways. (i) The first is to map the extracted vectors from the table to an arbitrary length through the fully-connected layer, and then use the broadcasting mechanism to diffuse them into a 3D feature map. We then concatenate the encoding tensor and fusion tensor. (ii) The second one maps the vectors extracted from the table to the channel lengths of the feature maps in the Bottleneck layer through the fully connected layer. After that, the feature tensor is expanded to the same dimension and size as the feature map using the broadcast mechanism as in the first method, and the expanded feature vector and feature map are directly summed. (iii) The third one is through Cross Attention mechanism, where the feature map is used as Key and Value, and the vector features are used as Query to calculate the attention weights. After comprehensive experiments, we find that the third solution using Cross Attention fusion ensures stable improvement.

To train the model, we need to select a suitable loss function. Considering that image segmentation is a pixel-level classification task, we explore the loss functions DiceLoss and HausdorffDTLoss as well as the cross-entropy loss and FocalLoss, which are common for classification. From comprehensive experiments, we find that combing the DiceLoss, FocalLoss, and cross-entropy as a new loss function can improve the model performance. Computing the HausdorffDTLoss takes too long time, and there is no significant performance improvement, so we abandon this loss in our final loss function.

We final loss function is provided below:

$$L = \alpha_1 L_{CE} + \alpha_2 L_{Dice} + \alpha_3 L_{Focal}$$

Where $\alpha_i$ is the weight of the three loss functions. The $L_{CE}$ and $L_{Dice}$[8] and $L_{Focal}$[9] formula is given by:

$$L_{CE} = -\frac{1}{N}2\sum_{i=1}^{N}\sum_{c=1}^{C} g_i^c \log s_i^c$$

$$L_{Dice} = 1 - \frac{2\sum_{i=1}^{N}\sum_{c=1}^{C} g_i^c s_i^c}{\sum_{i=1}^{N}\sum_{c=1}^{C} g_i^{c2} + \sum_{i=1}^{N}\sum_{c=1}^{C} s_i^{c2}}$$

$$L_{Focal} = \sum_{c=1}^{C} -\alpha_c(1-s_i^c)^\gamma \log(s_i^c)$$

Where $g_i^c$ is binary indicator if class label c is the correct classification for pixel i, and $s_i^c$ is the corresponding predicted probability.

$\alpha_c$ is a category-specific c weighting factor that is used to balance the loss contribution between different categories, thus helping to mitigate the category imbalance problem. $\gamma$ is the focusing parameter, which serves the same purpose as in the binary classification problem, and is used to reduce the loss contribution of the easy-to-categorise samples, making the model more focused on the difficult and misclassified samples. When $\gamma = 0$, the focal loss function is the standard cross entropy criterion.

### C. Post-Processing

To prevent the out-of-memory issue, we use a sliding-window strategy to reduce the size of the image of the chopping operation. We detect the size of the linkage area for each label after prediction, and for the abnormally small linkage area we merge it into the labels of the surrounding larger linkage area. This reduces the occurrence of voxel misclassification at the edges of the chunking.

### V. RESULT

The following is the final result of training the model under the nnU-net framework as published by the German Cancer Research Center. Since we use multiple servers to train, the reported results may differ from those using the MONAI framework in the Notebook.

The performance measures include Dice, HD95, MIoU, accuracy, recall, and precision. We first train U-Net, UNETER, SwinUNETR, MedNeXt, and our proposed Swin-NeXt. The formula is provided below.

1.Dice

$$\text{Dice} = \frac{2 \times |X \cap Y|}{\lceil X \rceil + \lceil Y \rceil}$$

where X is the predicted segmentation result, Y is the true segmentation result, $|X \cap Y|$ denotes the number of correctly predicted positive class pixel points, and $\lceil X \rceil$ and $\lceil Y \rceil$ are the number of predicted positive class pixel points and the true positive class pixel points, respectively.

2 Mean Intersection over Union

$$MIoU = \frac{1}{N}\sum_{i=1}^{N} \frac{|X_i \cap Y_i|}{|X_i \cup Y_i|}$$

where N is the number of categories, $X_i$ and $Y_i$ are the prediction results and true labels for the ith category, respectively, $|X_i \cap Y_i|$ is the number of pixel points that are correctly predicted, and $|X_i \cup Y_i|$ is the number of pixel points in the predicted and true labels belonging to the ith category.

3. Hausdorff Distance (HD95)

The Hausdorff distance is used as a measure of the maximum distance between two point sets, and HD95 is the 95th percentile of all point pairs in the Hausdorff distance, and is used to reduce the effect of outliers.

$$HD95(X, Y) = \max(h_{95}(X, Y), h_{95}(Y, X))$$
$$h_{95} = 95\text{th percentile of}\{\min_{y \in Y} d(x, y) : x \in X\}$$

where $d(x, y)$ is the Euclidean distance from point x to point y. X and Y are the sets of boundary points for the predicted and true segmentation results, respectively.

4. Accuracy Recall and Precision

$$ACC = \frac{TP + TN}{TP + TN + FP + FN}$$

$$Recall = \frac{TP}{TP + FN}$$

$$Precision = \frac{TP}{TP + FP}$$

where TP, TN, FP, and FN represent the number of pixel points in the true, true-negative, false-positive, and false-negative classes, respectively.

The detailed comparison is provided below.( It should be noted that our pre-defined experiments did not run out before the submission deadline, and the order in the table below represents the chronological order in which the models started to run. This is due to the queuing of tasks like server submissions. The table below only records the final run results of all models captured 1 hour before the submission deadline, and does not represent the real performance of the models):

| Model | Class | Dice | MIoU | HD95 | Accuracy | Recall | Precision |
|---|---|---|---|---|---|---|---|
| U-net | Liver | 0.860 | 0.780 | 4.613 | 0.989 | 0.870 | 0.855 |
|  | **Tumor** | **0.684** | **0.569** | 24.927 | **0.994** | **0.742** | **0.685** |
|  | vein | 0.528 | 0.395 | 15.644 | 0.999 | 0.550 | 0.554 |
|  | Aorta | 0.476 | 0.334 | 39.055 | 0.999 | 0.693 | 0.430 |
|  | Total | 0.709 | 0.614 | 16.847 | 0.991 | 0.770 | 0.704 |
| UNETR | Liver | 0.808 | 0.712 | 20.845 | 0.983 | 0.849 | 0.783 |
|  | Tumor | 0.468 | 0.344 | 63.410 | 0.987 | 0.507 | 0.539 |
|  | vein | 0.502 | 0.371 | 24.206 | 0.999 | 0.551 | 0.510 |
|  | Aorta | 0.456 | 0.324 | 46.932 | 0.999 | 0.616 | 0.420 |
|  | Total | 0.645 | 0.547 | 31.093 | 0.984 | 0.702 | 0.650 |
| SwinUNETR | Liver | 0.786 | 0.679 | 38.122 | 0.978 | 0.867 | 0.729 |
|  | Tumor | 0.502 | 0.364 | 93.55 | 0.988 | 0.584 | 0.502 |
|  | vein | 0.492 | 0.358 | 32.629 | 0.999 | 0.579 | 0.466 |
|  | Aorta | 0.508 | 0.366 | 22.201 | 0.999 | 0.649 | 0.509 |
|  | Total | 0.656 | 0.55 | 37.3 | 0.983 | 0.733 | 0.64 |
| MedNeXt | Liver | 0.852 | 0.761 | 4.148 | 0.988 | 0.859 | 0.849 |
|  | Tumor | 0.596 | 0.464 | 32.46 | 0.992 | 0.687 | 0.577 |
|  | vein | 0.503 | 0.37 | 21.382 | 0.999 | 0.514 | 0.543 |
|  | Aorta | 0.47 | 0.331 | 50.01 | 0.999 | 0.708 | 0.405 |
|  | Total | 0.683 | 0.583 | 21.6 | 0.99 | 0.752 | 0.674 |
| Swin-NeXt | Liver | 0.875 | 0.784 | 3.483 | 0.99 | 0.875 | 0.883 |

|       | Class | Dice  | MIoU  | HD95   | Accuracy | Recall | Precision |
|-------|-------|-------|-------|--------|----------|--------|-----------|
|       | Tumor | 0.613 | 0.490 | **22.165** | 0.993 | 0.624 | 0.726 |
|       | vein  | 0.431 | 0.303 | 17.973 | 0.993 | 0.452 | 0.480 |
|       | Aorta | 0.396 | 0.275 | 29.447 | 0.999 | 0.622 | 0.335 |
|       | Total | 0.662 | 0.569 | 14.614 | 0.992 | 0.714 | 0.684 |

Based on the above table, we propose to use the predicted values of nnU-net as the final results of this Challenge, because in the actual training, due to equipment constraints, nnU-net was run on the server for about 40 hours more than Swin-NeXt (twice the time and epoch), and thus obtained better results on the test set.

In order to tightly control the environmental variables, we collated the Dice on the test set for all models RTX4090 (24GB) running environment trained in 100 rounds and 24 hours respectively for comparison. The above comparison illustrates that our proposed Swin-NeXt and MedNeXt generally outperforms other methods in same epoch. The charts will be subsequently posted on Appendix in order to better show our comparison results.

We then test the three possible feature fusion methods proposed in Section III-B using our model and the two best-performing benchmarks. The results are summarized below. We can see that the fusion method using Cross Attention performs the best.

| Model | Class | Dice | MIoU | HD95 | Accuracy | Recall | Precision |
|-------|-------|------|------|------|----------|--------|-----------|
| SwinUNETR +Add | Liver | 0.819 | 0.724 | 13.48 | 0.982 | 0.853 | 0.81 |
| | Tumor | 0.49 | 0.36 | 73.565 | 0.988 | 0.522 | 0.544 |
| | vein | 0.503 | 0.372 | 19.795 | 0.999 | 0.592 | 0.482 |
| | Aorta | 0.507 | 0.371 | 27.94 | 0.999 | 0.604 | 0.521 |
| | Total | 0.662 | 0.562 | 26.956 | 0.987 | 0.713 | 0.67 |
| SwinUNETR +Concat | Liver | 0.79 | 0.686 | 34.08 | 0.978 | 0.868 | 0.743 |
| | Tumor | 0.429 | 0.301 | 114.394 | 0.986 | 0.536 | 0.458 |
| | vein | 0.477 | 0.343 | 33.151 | 0.999 | 0.592 | 0.446 |
| | Aorta | 0.518 | 0.377 | 14.289 | 0.999 | 0.608 | 0.539 |
| | Total | 0.641 | 0.537 | 39.197 | 0.981 | 0.717 | 0.636 |
| SwinUNETR +Cross Attention | Liver | 0.803 | 0.709 | 24.712 | 0.98 | 0.84 | 0.782 |
| | Tumor | 0.525 | 0.395 | 70.397 | 0.988 | 0.599 | 0.523 |
| | vein | 0.499 | 0.366 | 21.798 | 0.999 | 0.51 | 0.553 |
| | Aorta | 0.51 | 0.366 | 47.438 | 0.999 | 0.633 | 0.519 |
| | Total | 0.666 | 0.564 | 32.883 | 0.983 | 0.714 | 0.675 |
| MedNeXt +Add | Liver | 0 | 0 | 42.578 | 0.957 | 0 | 0.103 |
| | Tumor | 0 | 0 | 64.199 | 0.985 | 0 | 0.008 |
| | vein | 0 | 0 | 108.473 | 0.999 | 0 | 0 |
| | Aorta | 0.007 | 0.003 | 167.973 | 0.997 | 0.018 | 0.005 |
| | Total | 0.195 | 0.188 | 76.694 | 0.94 | 0.203 | 0.211 |
| MedNeXt +Concat | Liver | 0.401 | 0.255 | 113.6 | 0.899 | 0.774 | 0.275 |
| | Tumor | 0.009 | 0.004 | 231.91 | 0.955 | 0.018 | 0.01 |
| | vein | 0 | 0 | 225.17 | 0.998 | 0 | 0 |
| | Aorta | 0 | 0 | 81.27 | 0.999 | 0 | 0 |
| | Total | 0.268 | 0.226 | 130.547 | 0.878 | 0.335 | 0.254 |
| MedNeXt +Cross Attention | Liver | 0.862 | 0.782 | 4.289 | 0.989 | 0.864 | 0.863 |
| | Tumor | 0.573 | 0.457 | 57.272 | 0.992 | 0.663 | 0.549 |
| | vein | 0.516 | 0.382 | 19.836 | 0.999 | 0.499 | 0.585 |

| Model | Class | | | | | | |
|---|---|---|---|---|---|---|---|
| | Aorta | 0.468 | 0.332 | 50.743 | 0.999 | 0.635 | 0.464 |
| | Total | 0.683 | 0.589 | 26.428 | 0.99 | 0.731 | 0.692 |
| | Liver | 0.544 | 0.404 | 86.138 | 0.955 | 0.64 | 0.484 |
| Swin-NeXt +Add | Tumor | 0 | 0 | 96.91 | 0.985 | 0 | 0 |
| | vein | 0 | 0 | 95.769 | 0.999 | 0 | 0 |
| | Aorta | 0 | 0 | 189.228 | 0.999 | 0 | 0 |
| | Total | 0.304 | 0.271 | 93.623 | 0.955 | 0.323 | 0.292 |
| | Liver | 0.542 | 0.388 | 88.745 | 0.948 | 0.716 | 0.448 |
| Swin-NeXt +Concat | Tumor | 0 | 0 | 42.898 | 0.985 | 0 | 0.041 |
| | vein | 0 | 0 | 118.105 | 0.999 | 0 | 0 |
| | Aorta | 0 | 0 | 117.414 | 0.999 | 0 | 0 |
| | Total | 0.303 | 0.267 | 73.432 | 0.953 | 0.336 | 0.294 |
| | Liver | 0.879 | 0.790 | 2.830 | 0.991 | 0.873 | 0.892 |
| Swin-NeXt +Cross Attention | Tumor | **0.620** | **0.500** | **13.331** | **0.993** | **0.642** | **0.704** |
| | vein | 0.430 | 0.302 | 18.076 | 0.999 | 0.442 | 0.491 |
| | Aorta | 0.361 | 0.241 | 22.164 | 0.999 | 0.598 | 0.309 |
| | Total | 0.657 | 0.565 | 11.28 | 0.991 | 0.710 | 0.678 |

Another interesting finding we made based on tracing the experimental process is that the Swin-NeXt and MedNeXt models do not converge when performing the two feature fusion methods, Add and Concat, whereas SwinUNETR converges well; the most notable structural difference between the three is the fact that the former two use a transposed convolution for the up-sampling in the decoder, while the latter uses an upper adoption layer to accomplish this operation. Therefore, we can assume that the transposed convolution structure is not suitable for decoding feature maps based on additive feature fusion via Add and Concat.

On the contrary, Cross Attention can give a stable boost to the original model.

Finally, we chose CT slices from patient HCC066 as a comparison and give below a few visualisations of the segmentation results for the better performing models.
We can find that although our model does not perform numerically up to SOTA, it is the closest to the real labels in terms of both the size and shape of the segmentation, so our work may be more informative.

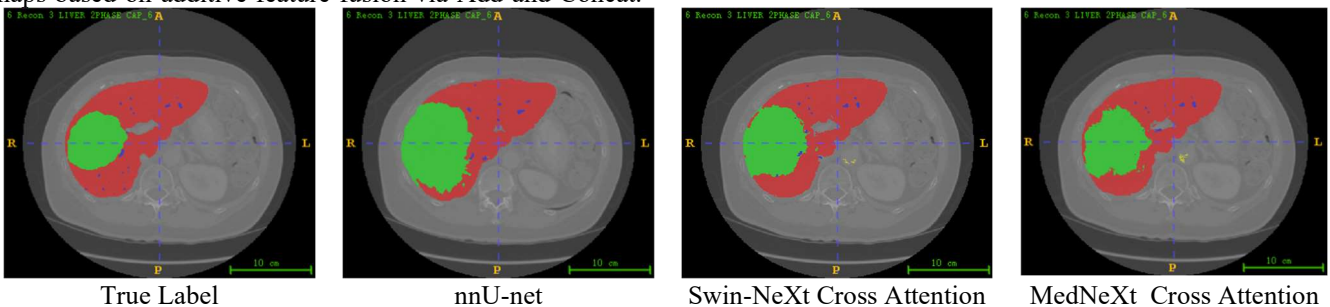

| True Label | nnU-net | Swin-NeXt Cross Attention | MedNeXt Cross Attention |

## VI. CONCLUSION AND REFLECTION

Our work has the following contributions:

First, we combined the encoder part of SwinUNETR and the decoder part of MedNeXt and constructed a cross-attention module to perform modal fusion on the last layer of the SwinUNETR encoder, which enables the model to accept the pathology and treatment information of the samples in vector format to assist the segmentation task during training and inference. We found that the Transformer-based Swin Transformer has the strongest ability to encode 3D images, the convolution-based Conv-NeXt module has the strongest ability to decode at the pixel level, and the cross-attention modal fusion has the best results.

Second, we combined and simplified the massive existing pre-processing post-processing as well as data enhancement methods in 3D image segmentation based on dynUnet[10] and nnU-Net frameworks with brief ablation experiments. We integrated our proposed Swin-NeXt with Cross-Attention framework into this framework, which makes the model converge faster. We also modified the loss function by splitting the DiceCELoss function into DiceLoss and cross-entropy loss and adding FocalLoss, which improves the convergence speed of the model on datasets with imbalanced voxel labels.

Meanwhile, there is something to reflect on. Later in the competition, we actively reduced the number of parameters in our model in order to compare it more fairly with other models, and abandoned the network structure with the optimal feature extraction block ratio of 1:1:3:1, which had been demonstrated in both Swin transformer and ConvNeXt[11], which was the possible reason why we did not exceed the scores of the SOTA.

ACKNOWLEDGMENT

We would like to thank the 2024 IISE DAIS Case Study Competition Committee that organized the competition and labeled the raw data.

APPENDIX

| Model | 100 Epoch | 24 Hour |
|---|---|---|
| UNETR | 0.43 | 0.52 |
| nnU-net | **0.48** | 0.55 |
| SwinUNETR | 0.47 | 0.52 |
| MedNeXt | 0.44 | 0.50 |
| **Swin-NeXt** | 0.44 | 0.54 |
| **SwinUNETR +Add** | 0.45 | **0.56** |
| **SwinUNETR +Concat** | 0.44 | 0.53 |
| **SwinUNETR +Cross Attention** | 0.45 | 0.53 |
| MedNeXt +Add | — | — |
| MedNeXt +Concat | — | — |
| **MedNeXt +Cross Attention** | 0.45 | 0.54 |
| Swin-NeXt +Add | — | — |
| Swin-NeXt +Concat | — | — |
| **Swin-NeXt +Cross Attention** | 0.45 | **0.56** |